\documentclass[letterpaper, 10 pt, conference]{ieeeconf}  

\IEEEoverridecommandlockouts                              

\usepackage{cite}
\usepackage{amsmath,amssymb,amsfonts}
\usepackage{algorithmic}
\usepackage{graphicx}
\usepackage{textcomp}
\usepackage{xcolor}
\usepackage{cite}
\usepackage{dirtytalk}
\usepackage{hyperref} 
\usepackage{diagbox}
\usepackage{color,soul}
\usepackage{flushend} 

\usepackage{subfigure}
\usepackage{tikz}

\definecolor{red}{rgb}{1.00,0.00,0.00}
\definecolor{blue}{rgb}{0.00,0.00,1.00}
\definecolor{green}{rgb}{0.30, 0.50,0.00}
\newcommand{\cred}[1] {\textcolor{red}{#1}}
\newcommand{\cblue}[1] {\textcolor{blue}{#1}}

\usepackage{lipsum}
\def\BibTeX{{\rm B\kern-.05em{\sc i\kern-.025em b}\kern-.08em
    T\kern-.1667em\lower.7ex\hbox{E}\kern-.125emX}}
    
\title{\LARGE \bf
Self-supervised Learning for Joint Pushing and Grasping Policies in Highly Cluttered Environments
}

\author{Yongliang Wang$^1$\textsuperscript{*}, Kamal Mokhtar$^1$\textsuperscript{*}, Cock Heemskerk$^2$, Hamidreza Kasaei$^1$ 
\thanks{* These authors contributed equally to this work. \newline
${^1}$Department of Artificial Intelligence, Bernoulli Institute, University of Groningen, 9747 AG, The Netherlands. \newline 
	${^2}$ Heemskerk Innovative Technology, Delft, The Netherlands\newline
	 email: yongliang.wang@rug.nl,
 mokhtar.kamal@icloud.com,\newline
 hamidreza.kasaei@rug.nl}
}

\begin{document}
\maketitle
\thispagestyle{empty}
\pagestyle{empty}

\begin{abstract}
Robotic systems often face challenges when attempting to grasp a target object due to interference from surrounding items. We propose a Deep Reinforcement Learning (DRL) method that develops joint policies for grasping and pushing, enabling effective manipulation of target objects within untrained, densely cluttered environments. In particular, a dual RL model is introduced, which presents high resilience in handling complicated scenes, reaching an average of $98\%$ task completion in simulation and real-world scenes. To evaluate the proposed method, we conduct comprehensive simulation experiments in three distinct environments: densely packed building blocks, randomly positioned building blocks, and common household objects. Further, real-world tests are conducted using actual robots to confirm the robustness of our approach in various untrained and highly cluttered environments. The results from experiments underscore the superior efficacy of our method in both simulated and real-world scenarios, outperforming recent state-of-the-art methods. To ensure reproducibility and further the academic discourse, we make available a demonstration video, the trained models, and the source code for public access. \href{https://github.com/Kamalnl92/Self-Supervised-Learning-for-pushing-and-grasping}{\cblue{https://github.com/Kamalnl92/Self-Supervised-Learning-for-pushing-and-grasping}}.

\end{abstract}

\section{Introduction}

Grasping is fundamental to a myriad of robotic applications, facilitating tasks that are untenable without adept object manipulation~\cite{newbury2023deep}. In many situations, objects are not isolated, making grasping challenging due to environmental occlusions and clutter~\cite{hoang2022context, zuo2023graph, lu2023picking}. Drawing inspiration from human dexterity, robots should employ motion primitives, like pushing, to isolate target objects from clutter and facilitate grasping~\cite{huang2022estimating, kang2022grasp}. For effective grasping, robots need to integrate visual observations of the goal object(s) and understand spatial relationships, rather than just executing a basic antipodal grasp. This necessitates a harmonious system where all components collaborate~\cite{burgess2022dgbench}. Given the potentially flawed sensory information a robot obtains from its environment, it needs to effectively address these inconsistencies.
Additionally, dynamic scene changes after each action necessitate the adaptability of the robot before task completion~\cite{duan2023semantic}. Moreover, planning the manipulator's motion to achieve the desired pose is inherently challenging, requiring continuous environment observation and adjustments~\cite{kunz2010real, wu2023efficient, wu2023learning, zhou2023learning, zhang2023reinforcement, lu2023picking}.

\begin{figure}[!t]
    \centering
    \includegraphics[width=1.0\columnwidth, trim= 0cm 0mm 0cm 1mm, clip=true]{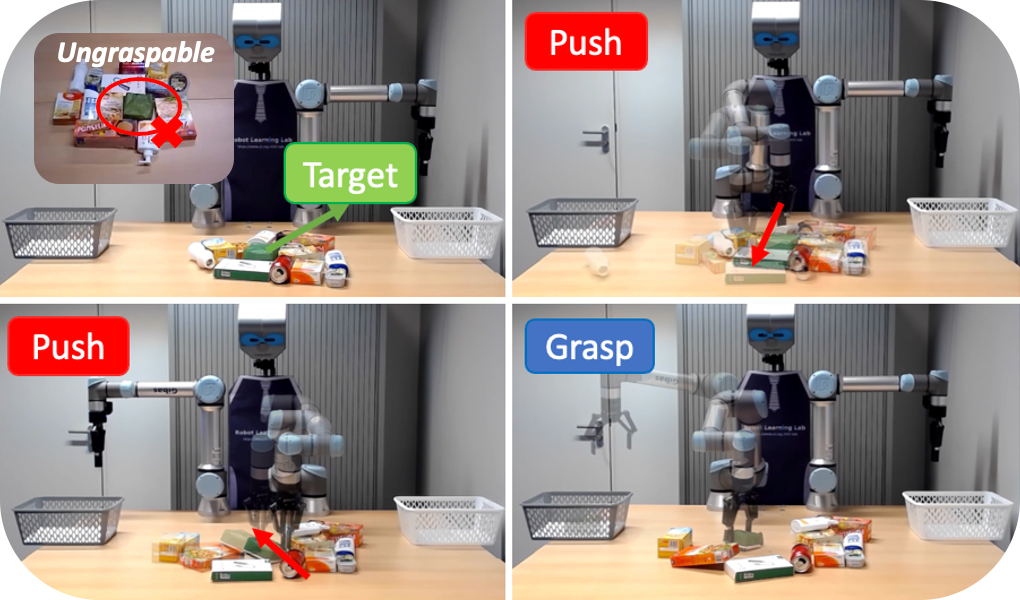}
    \caption{Achieving Grasping Amidst Highly Cluttered Objects: The green target object, initially ungraspable due to surrounding blocks, is repositioned using pre-manipulation pushes by the robot. This strategic maneuver ensures a feasible grasp, showcasing the efficacy of our method.}
    \label{fig:PushGrasp}
    \vspace{-3mm}
\end{figure}

This paper primarily addresses self-supervised learning for pushing and grasping in cluttered settings through DRL~\cite{fujita2020important, xu2021efficient, berscheid2019robot}. As depicted in Fig.~\ref{fig:PushGrasp}, when a direct grasp of the target object is unfeasible, our method employs a combination of pushing and grasping to guarantee successful manipulation. A multitude of studies have delved into the conjunction of pushing and grasping, probing their combined impact on object manipulation~\cite{murali20206, tang2021learning, yang2020deep}.
While most existing studies focus on bin-picking tasks, which entail transferring objects between bins, our research investigates situations where surrounding clutter renders the target object initially ungraspable. The interplay between pushing and grasping cultivates a unified behavior, elevating the proficiency of object manipulation~\cite{huang2017densely, liu2022ge}. Instead of considering pushing and grasping as separate actions, which can result in unforeseen scene alterations and problem-solving hurdles, our method utilizes DRL to harness the power of synergy and address its limitations. The complete system architecture is depicted in Fig.~\ref{fig:wholeSystem} and detailed in Section \ref{sec:method}. Broadly, our paper provides the following contributions:

\begin{figure*}[h!]
    \centering
    \includegraphics[width=1\textwidth]{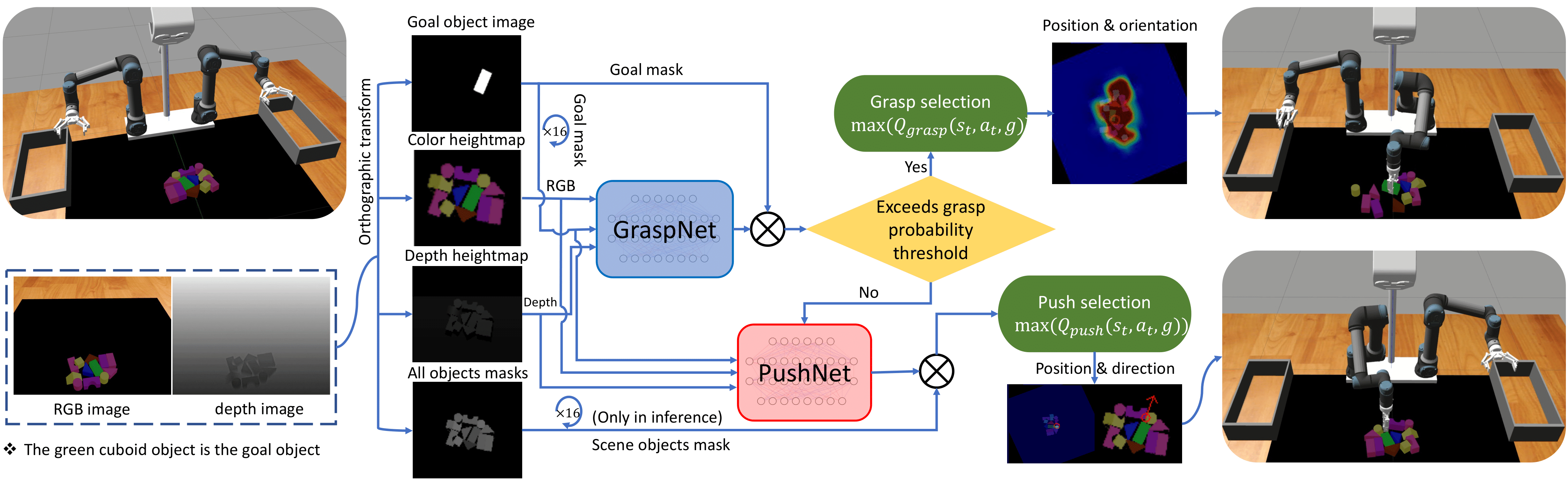}
    \caption{\textbf{System Overview}: Within the Gazebo environment, an RGB-D camera and dual-arm UR5e robot are integrated. The camera transforms sensory data into orthographic projections, generating goal and scene masks. Using 360° rotation, the grasp and push nets assess heightmaps and the goal mask, outputting Q values to dictate push or grasp actions.}
    \label{fig:wholeSystem}
\end{figure*}

\begin{itemize}
    \item We propose a self-supervised learning method enabling a service robot to concurrently learn pushing and grasping policies, aiming to efficiently clear occlusions around target objects. Our approach addresses intricate tasks like obstacle removal while probing the constraints of the model and system.
    \item We conduct extensive simulations and real-world experiments to validate our method. Simulations encompass scenarios with building blocks in both packed and random arrangements. Empirical studies using the real robot further affirm the robustness of our approach. The results convincingly outperform current state-of-the-art strategies in task completion and grasp success rates, detailed in Section~\ref{sec:Experiment}.
    \item For the broader research community's benefit, we make our source code and trained models publicly accessible, facilitating result replication.\footnote{\href{https://github.com/Kamalnl92/Self-Supervised-Learning-for-pushing-and-grasping}{\resizebox{0.95\linewidth}{!}{https://github.com/Kamalnl92/Self-Supervised-Learning-for-pushing-and-grasping}}}.
\end{itemize}

\section{Related Work}

In recent years, advancements in robotic grasping have been noteworthy. This section explores object grasping and pre-grasp manipulation~\cite{wu2023learning, ghorbani2023dual}.

\subsection{Grasping}
Autonomous grasping has advanced rapidly, becoming foundational for robotic tasks in complex environments~\cite{duan2021robotics}. While traditional methods depended on precise object modeling for grasp metrics, they often faced impractical demands~\cite{sahbani2012overview}. Ensuring stable interactions between the robot's hand and various objects necessitated accurate mathematical representations~\cite{prattichizzo2016grasping}. With recent computational advancements, the paradigm has shifted to deep-learning and data-driven methods. Leveraging image and depth data, these methods enable robots to simulate and perform grasps without exhaustive object knowledge, overcoming previous constraints~\cite{duan2021robotics}.

Data-driven grasping methods prioritize the analysis of successful grasp instances over object mechanics. By utilizing grasp simulators, these approaches assess grasp stability based on manually gathered or simulated 3D environment data~\cite{miller2003automatic, goldfeder2007grasp, ciocarlie2007dexterous}. The rise of geometry-based antipodal grasp evaluation has invigorated the use of deep learning in this field~\cite{pas2015using, ten2017grasp}, with innovations like the GR-ConvNet~\cite{kumra2020antipodal}. The emergence of multi-view 3D grasping methods, such as those representing objects through three perspective depth images, provides grasp heatmaps~\cite{kasaei2021mvgrasp}. While grasp simulators are adept at evaluating known objects, addressing unknown ones involves shape completion prior to grasping~\cite{varley2017shape, van2020learning} or image-based grasp pose inference~\cite{lenz2015deep, mahler2017dex, schmidt2018grasping, lu2020multi}. Current methods typically concentrate on isolated objects in uncluttered environments. For cluttered scenarios, most solutions employ parallel grippers. Mousavian et al. proposed a grasp-generating method through a variational autoencoder, circumventing random sampling~\cite{mousavian20196}. However, antipodal-centric approaches might falter in cluttered areas necessitating precise grasping~\cite{xu2021adagrasp, wu2020generative}. To address these challenges, Kiatos et al. proposed a system that combines scene point clouds with robot hand geometry. However, this system encounters difficulties in achieving collision-free grasps in restricted spaces~\cite{kiatos2020geometric}.

\subsection{Pre-grasp manipulation}
In cluttered environments, pre-grasp manipulations are crucial for effective grasps. While RL excels in isolating objects from obstructions~\cite{mohammed2020review}, a research gap exists in extracting specific objects from such environments, essential for service robots. Successful grasping often requires manipulating surrounding objects. Within RL, integrating a Markov Decision Process (MDP) is fundamental. However, the intrinsic data-heavy nature of RL, as well as the issue of scarce rewards in time-bound tasks, presents difficulties. An adjustment in the MDP, tying state-to-action and refining skill allocation in grasping, offers a solution~\cite{berscheid2019robot}. Despite the prevalence of top-down grasping approaches, the diverse orientations of objects necessitate a range of grasp evaluations. One method involves an initial learning phase from isolated items, followed by a discriminative model that considers possible collisions. Following that, a Variational Autoencoder (VAE) is used to evaluate grasps and validate collision outcomes.~\cite{murali20206}.

Visual Pushing for Grasping (VPG) proposed by Zeng et al. emphasized experiential learning, enabling robots to adapt their grasp based on object orientation from 16 potential push directions~\cite{zeng2018learning}. Bao et al. employed deep learning for task segmentation, though their reliance on color was restrictive~\cite{bao2022learn}. Ewerton et al. introduced an algorithm optimized for cylindrical objects, aiming to reduce push actions~\cite{vieira2022persistent}. Baris et al. leveraged deep Q-learning in cluttered settings~\cite{serhan2022push}, while Guo et al.'s graph-based DRL targeted occluded items but had a limited application range~\cite{zuo2022graph}.

While much of the existing research focuses on identifying objects in cluttered environments, our study extends to both densely packed and sporadically distributed object contexts. In highly cluttered environments, the efficient interplay of pushing and grasping remains challenging when success-based rewards are the only incentives. To enhance efficiency, we integrate Hindsight Experience Replay (HER) for goal relabeling, enriching the replay buffer, and accelerating learning~\cite{andrychowicz2017hindsight}. Building upon Xu et al.~\cite{xu2021efficient}, our paper further probes the push-grasp interrelation. By framing it as a MDP, a nuanced balance between deterministic and stochastic actions is maintained in our approach.

\section{Method}
\label{sec:method}
\subsection{Overall system}

As depicted in Fig.~\ref{fig:wholeSystem}, our system chooses between two actions: push and grasp, depending on environmental conditions. If grasp likelihood surpasses a threshold, it opts to grasp; otherwise, it pushes the target or adjacent objects approximately $1.3\,cm$ to enhance subsequent grasp opportunities. Grasp/push coordinates, initially in the image domain, are later converted to the robot's frame for action.

In our RL framework, we address pushing and grasping in highly cluttered environments. The policy, reward, and $Q$-value function are denoted by $\pi(s_t|g)$, $R(s_t,a_t,g)$, and $Q(s_t,a_t,g)$, respectively. Capturing the scene, a camera produces orthographic images which are rotated 16 times, each at 22.5 degrees, cumulatively spanning 360 degrees. This rotation facilitates the learning of grasp and push orientations. The action chosen aligns with the highest $Q$-value rotation. If this value exceeds $1.8$, an object grasp is pursued. Otherwise, a push action is employed, as elaborated by Xu et al. (2021)~\cite{xu2021efficient}. Mirroring Generative Adversarial Networks (GAN) principles, the grasp network $\phi_g$ operates as a discriminator, while the push network $\phi_p$ acts akin to a generator. $\phi_p$ modifies the scene to optimize the grasp's $Q$-value, centering on $Q_{grasp}$ (refer Fig.~\ref{fig:wholeSystem}). Our model selectively processes the target object's discrete mask, isolating relevant scene features.

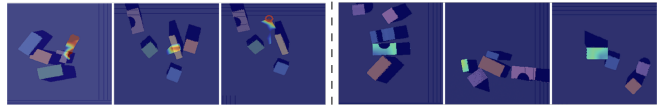
\begin{figure}[!t]
    \includegraphics[width=0.49\linewidth]{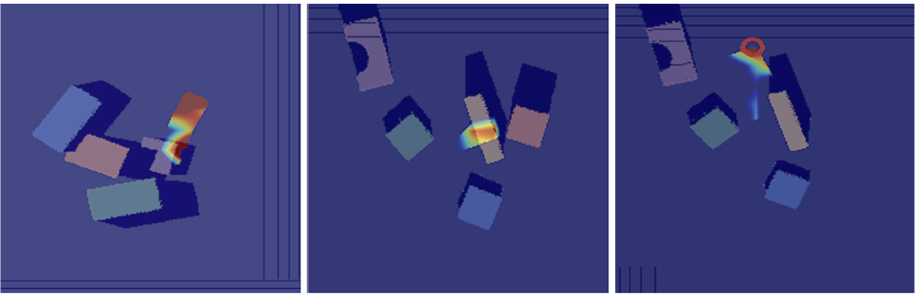}
    \begin{tikzpicture}
        \draw[dashed, line width = 0.01pt] (0,0) -- (0,0.16\linewidth);
    \end{tikzpicture}
    \includegraphics[width=0.49\linewidth]{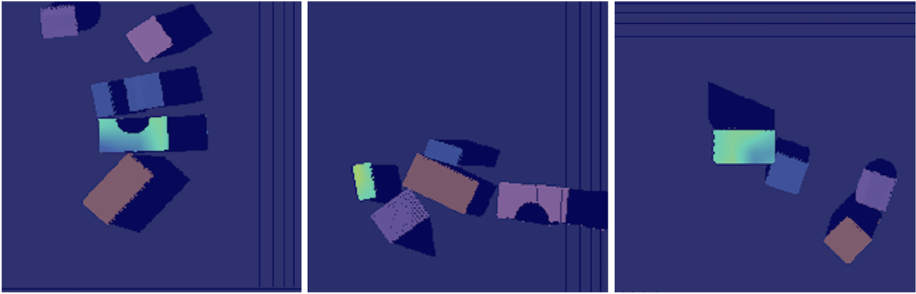}
\caption{Post-mask Pixel $Q$-values: (\textit{left}) Xu et al. (2021) flaws~\cite{xu2021efficient}; (\textit{right}) our refined method.}
\label{fig:masking}
\end{figure}

Our method diverges from Xu et al.(2021)~\cite{xu2021efficient}, who assign specific masks to each object. Rather than restricting our model to $Q$-values of relevant objects during training, we expand its exploration domain. The model assesses all $Q$-values, acting on the peak value. During inference, we overlay the object mask to guide actions. The discrete masking method by~\cite{xu2021efficient} can lead to erroneous outcomes. As illustrated at the left of Fig.\ref{fig:masking}, high pixel-wise Q-values in vacant areas might misdirect robotic tasks. This inconsistency, highlighted by the red circle, causes unintended maneuvers. Their strategy also necessitates different masking for training and testing, posing scalability challenges. Advanced segmentation methods, like those by Xiang et al. (2020)~\cite{xiang2020learning}, provide solutions. However, with our primary focus away from perception, we employ RGB-color segmentation, recognizing the target's green color. The effectiveness of our masking is evident at the right of Fig.~\ref{fig:masking}, following its application to pixel $Q$-values.

\subsection{Models hierarchical framework}
Our architecture features two distinct networks: the grasp ($\phi_g$) and push ($\phi_p$) networks, setting it apart from Xu et al. (2021)~\cite{xu2021efficient}. We employ dual 121-layer DenseNets~\cite{huang2017densely}, previously trained on ImageNet~\cite{deng2009imagenet}, reducing our model's scale to one-third of that presented by Xu et al. These DenseNets independently process RGB inputs and normalized depth images. Their resulting outputs merge into a cohesive feature vector, channeled into a fully convolutional layer~\cite{long2015fully} equipped with $1 \times 1$ kernels, ReLU activation functions, and batch normalization~\cite{ioffe2015batch}. Finally, bilinear upsampling is applied to ensure the output's dimensions align with the input.

\subsection{Models training}

\begin{figure}[!t]
    \includegraphics[width=\columnwidth] {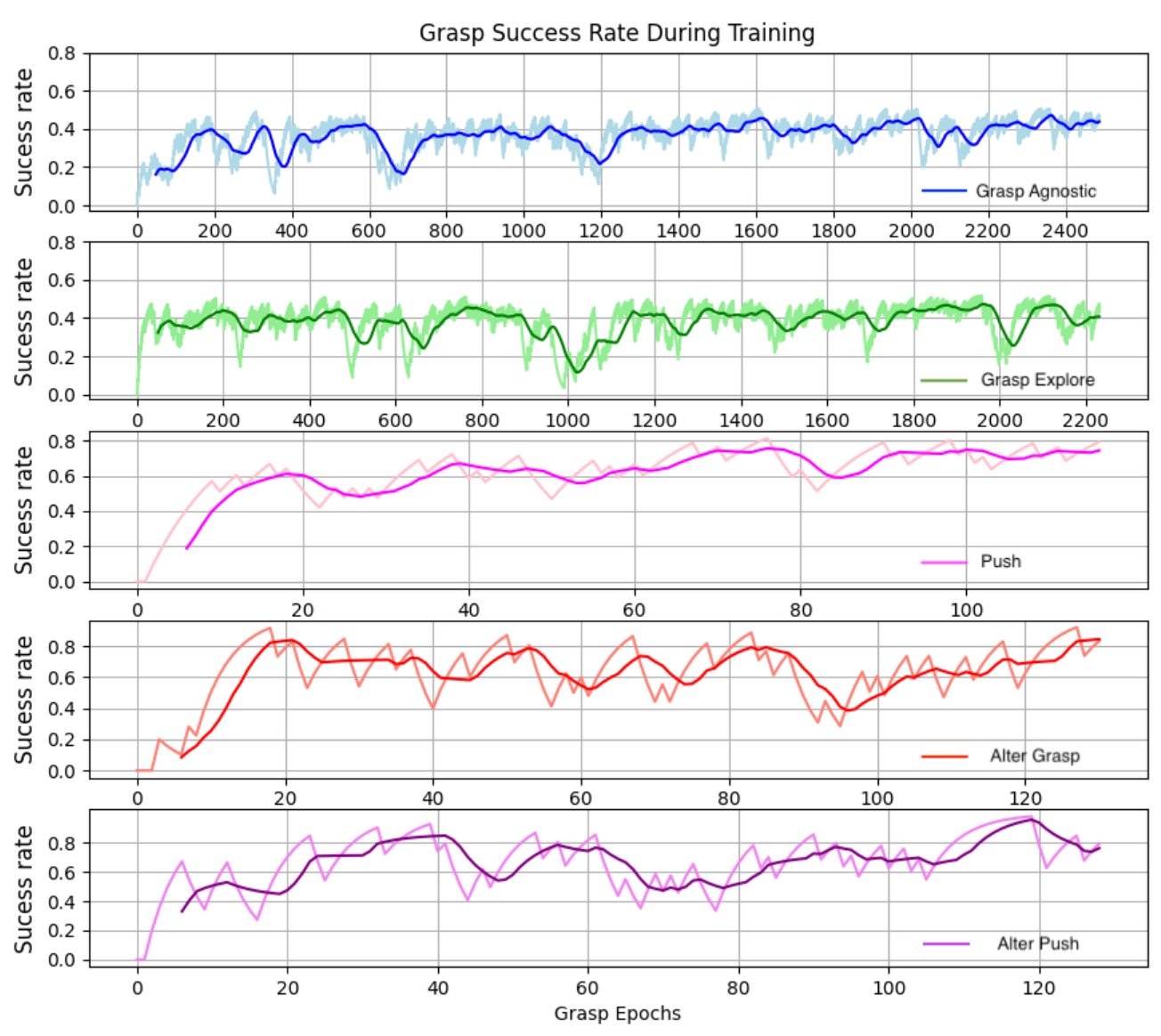}

    \caption{Grasp success rate versus the number of grasp epochs. The top two are related to the grasping network $\phi_g$ training, and the middle one is for the push network $\phi_p$ training. The last two are when the training is alternated between the grasp net and push net.}
    \label{fig:trainingresults}
\end{figure}

Our training process unfolds in three stages: starting with goal-conditioned grasping, transitioning to goal-conditioned pushing, and concluding with alternating training between both. The rewards for grasping, $R_g$, and pushing, $R_p$, are defined as follows:
\begin{equation}
    R_g = 
    \begin{cases}
        1, & \text{if grasp success}\\
        0, & \text{if not}
    \end{cases}
    \label{eq:greward}
\end{equation}
\begin{equation}
    R_p = 
    \begin{cases}
        0.5, & Q_g^{improved} > 0.1 \text{ \& scene change}\\
        -0.5, & \text{ no scene change}\\
        0, & \text{otherwise}
    \end{cases}
    \label{eq:preward}
\end{equation}
\noindent where $Q_g^{improved}$ is calculated as 
\begin{equation}
     Q_g^{improved} = Q_g^{post-push} - Q_g^{pre-push} 
    \label{eq:rewardImproved}
\end{equation}
Furthermore, changes in the scene are gauged from the depth map around the target object. Drawing inspiration from the Bellman equation in RL, we define the state-action function relative to the goal as $\pi(s|a,g)$. Employing the epsilon-greedy action selection, denoted as $\epsilon(\pi(s|a,g))$, allows the agent to balance exploration and exploitation~\cite{gimelfarb2020epsilon}.
\begin{equation}
    \nu(S_t,g) = E[R_{[t+1]}+\rho\nu(S_{[t+1]})|S_t=S, g] 
    \label{eq:eqBill}
\end{equation}
Where $\rho$ represents a discount function, an overview of the training process and the grasp success rate for the different stages of training are depicted in Fig.~\ref{fig:trainingresults}. 
In the following sub-sections, more details about the training are discussed.

\subsubsection{Goal-conditioned grasping}
During training, we set five objects in a sparse workspace, as depicted in Fig.~\ref{fig:masking}. Training is episodic, with each grasp marking an episode through two phases. In the \textit{grasp agnostic} phase, we use target relabeling to optimize sampling. Due to the better gripping ability of the model, we transition to the \textit{grasp explores} phase, eschewing goal relabeling, assuming the model efficiently gauges orientation. Both stages utilize an $\epsilon$-greedy strategy to balance exploration and exploitation~\cite{d2019exploiting}. Unlike Xu et al. (2021)~\cite{xu2021efficient}, who terminate training based on $Q$-values, we rely on grasp success rates. As evident in Fig.~\ref{fig:trainingresults}, training plateaus after 1400 epochs. It should be noted that, during the \textit{grasp explores} phase, despite a consistent success rate, the task's complexity heightens, given only specific grasps are deemed successful, more so with an unpolished push model.

\subsubsection{Goal-conditioned pushing}
In this training phase, we fix the grasp model and exclusively refine the push model through adversarial training. Each episode encompasses up to five pushes, culminating in a grasp. If the target object's grasp $Q$-value exceeds a certain threshold, the episode terminates with an immediate grasp by the robot. Within a mere 120 epochs, the push model's scene adjustments elevate the goal-directed grasp success rate significantly (see Fig.~\ref{fig:trainingresults}, third row). Rewards are allotted to the push model solely when it heightens the future grasp probability for the goal object.

\subsubsection{Goal-conditioned alternating}
Initially, the grasp model trains in sparse environments with 5-12 objects to minimize occlusion, leading to a distribution disparity. To reconcile this, after push model training, we fine-tune the grasp model. Both models then train collaboratively in 10-object scenarios for optimal coordination. Recognizing the erratic behavior of certain objects, like bottles, when pushed, we devise an alternative: the model prioritizes grasping non-target items to clear a path to the primary object, sequentially engaging with these obstacles until the main target becomes accessible.

\begin{figure*}[!t]
    \includegraphics[width=0.195\linewidth]{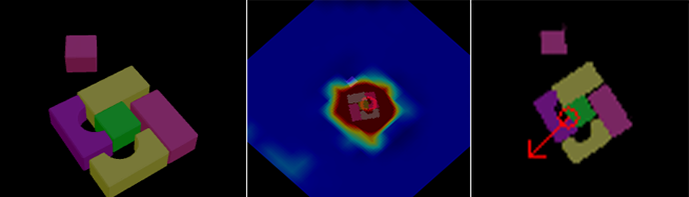} \hfill
    \includegraphics[width=0.195\linewidth]{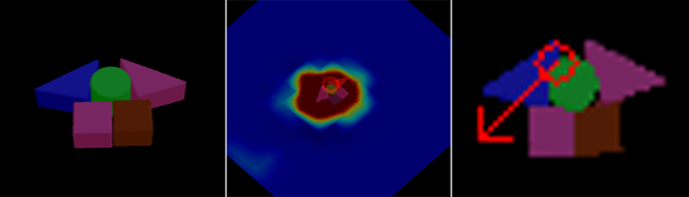} \hfill
    \includegraphics[width=0.195\linewidth]{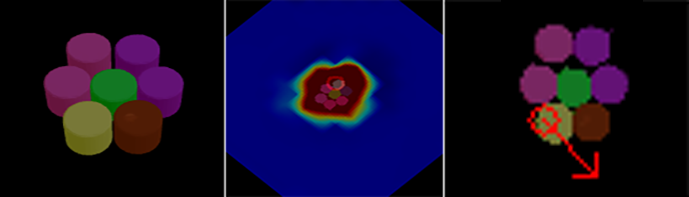} \hfill
    \includegraphics[width=0.195\linewidth]{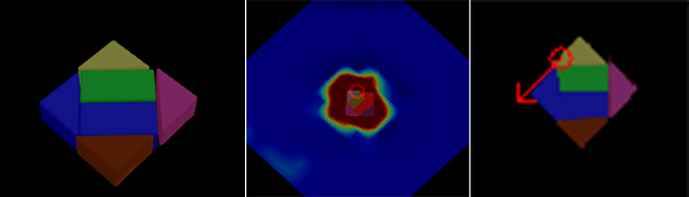} \hfill
    \includegraphics[width=0.195\linewidth]{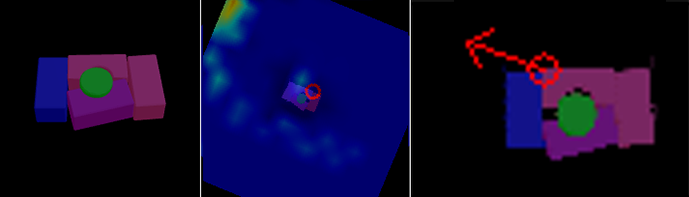}
    
    \par\noindent\makebox[0.195\linewidth][c]{\footnotesize(case:1)} \hfill
    \makebox[0.195\linewidth][c]{\footnotesize(case:2)} \hfill
    \makebox[0.195\linewidth][c]{\footnotesize(case:3)} \hfill
    \makebox[0.195\linewidth][c]{\footnotesize(case:4)} \hfill
    \makebox[0.195\linewidth][c]{\footnotesize(case:5)}

    \includegraphics[width=0.195\linewidth]{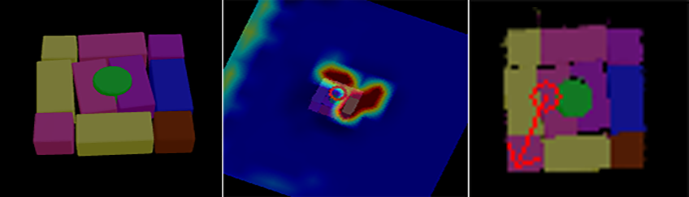} \hfill
    \includegraphics[width=0.195\linewidth]{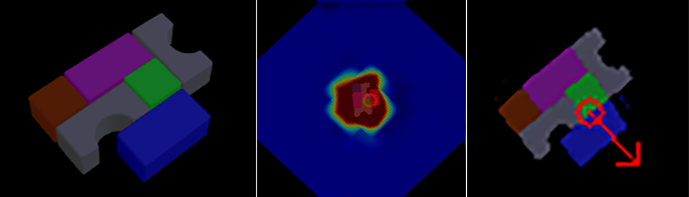} \hfill
    \includegraphics[width=0.195\linewidth]{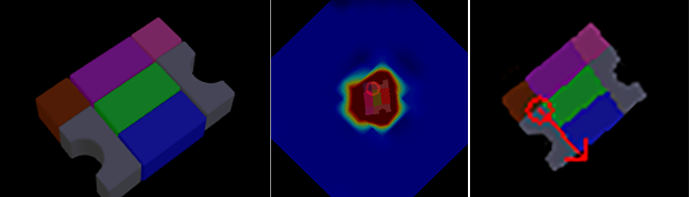} \hfill
    \includegraphics[width=0.195\linewidth]{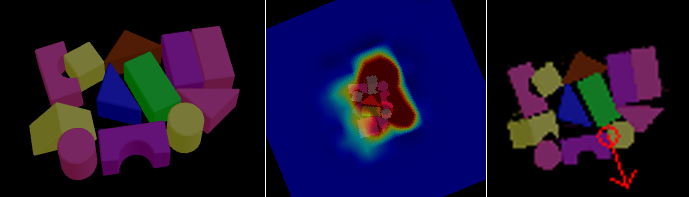} \hfill
    \includegraphics[width=0.195\linewidth]{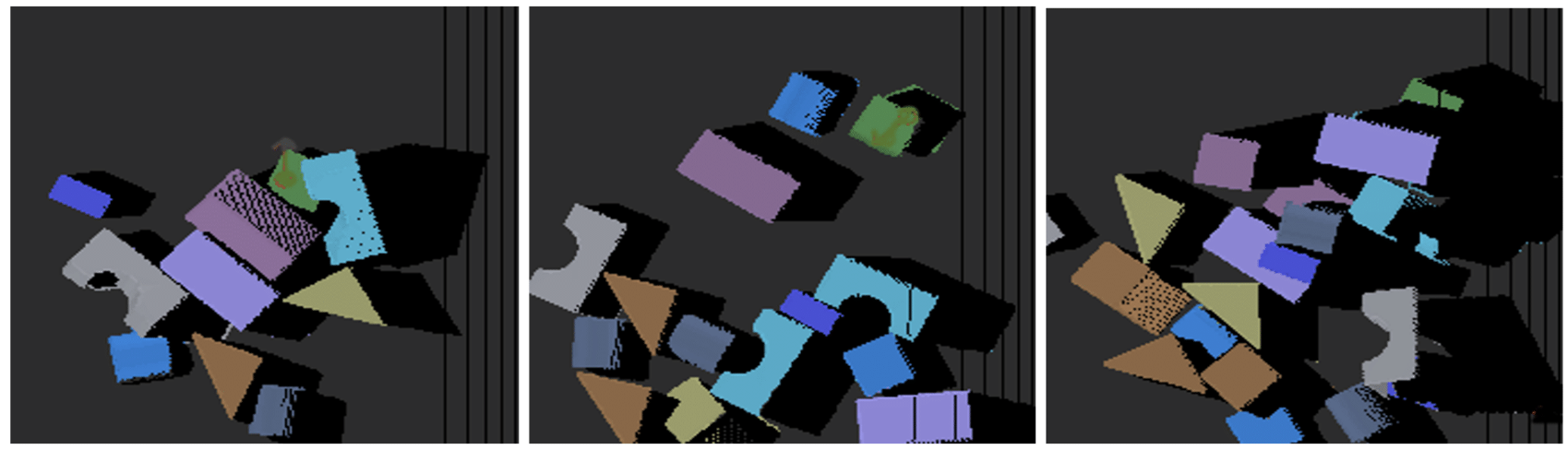}

    \par\noindent\makebox[0.195\linewidth][c]{\footnotesize(case:6)} \hfill
    \makebox[0.195\linewidth][c]{\footnotesize(case:7)} \hfill
    \makebox[0.195\linewidth][c]{\footnotesize(case:8)} \hfill
    \makebox[0.195\linewidth][c]{\footnotesize(case:9)} \hfill
    \makebox[0.195\linewidth][c]{\footnotesize(case:10)}

\caption{The first column shows the original image; the next depicts the output angle and position; the last highlights the initial action. Case $1-9$: Simulation experiments with 9 distinct packed scenes featuring dense adversarial clutter; each scene's target is the green object. Case $10$: Typical scenes showcasing random $10$, $15$, and $20$ objects from left to right.}
\label{fig:5}
\end{figure*}

\begin{figure}[!t]
    \includegraphics[width=1.0\linewidth]{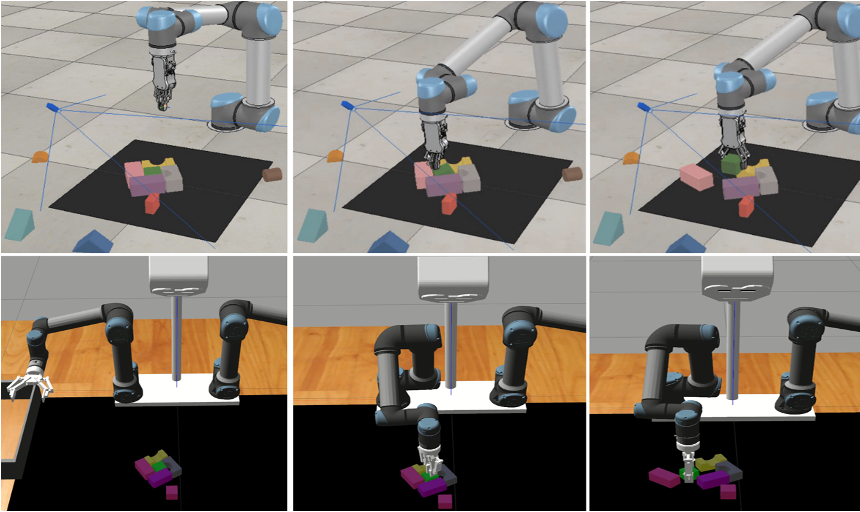}
\caption{Snapshot Sequence: Robot declutters for target grasp: The sequence includes (\textit{left}) the initial scene; (\textit{middle}) the process of pushing surrounding objects; (\textit{right}) the final act of grasping the goal object. Top: Coppeliasim; Bottom: Gazebo. Model trained in Coppeliasim, fine-tuned in Gazebo.}
\label{fig:packedExps}
\end{figure}

Our objective is to grasp a designated target, highlighted in green in Fig.~\ref{fig:5}, without disturbing nearby items. Instead of conventional multi-predictive methods, we determine the target using aggregated pixel-wise Q-values, which inform both its orientation and position. After training, challenges persisted near target areas, with success rates plateauing at $40\%$. This approach does not work for such cases. The consistent collision of the robot arm with objects in the scene and the sequence decoding of removing collision objects seem impossible for the model to learn. Approaches such as curriculum learning could mitigate the issue~\cite{narvekar2020curriculum}. It addresses a complex problem where a sequence of tasks is presented with an increase in difficulty.

\section{Experiment}
\label{sec:Experiment}
To assess our push-grasping strategy, we conducted tests in both simulated and real-world scenarios. The objective was to compare its efficiency with established policies and verify its stability in real robotic systems, with minimal reduction in performance.

\subsection{Simulation Experiments}
\subsubsection{Experimental Setups and Metrics}
Extensive experiments were finished utilizing the proposed method on the Coppeliasim and Gazebo simulators. We trained on Coppeliasim~\cite{CoppeliaSim} and conducted preliminary block-building tests to guarantee a fair comparison for both strategies. Subsequent Gazebo experiments delved into scenarios mimicking real-world conditions with common colored objects. The model trained on Coppeliasim was fine-tuned and deployed in Gazebo, allowing for the handling of real-world objects. The whole system, depicted in Fig.~\ref{fig:wholeSystem}, features dual UR5e robot arms and an RGB-D Intel RealSense SR300. For motion mapping, we employed an inverse kinematics (IK) solver~\cite{diankov2010automated}, and for efficient training, the networks used the Adam optimizer~\cite{kingma2014adam} on an NVIDIA V100 GPU. For the valid comparison, we compared our method with Xu et al.~\cite{xu2021efficient}. As far as we know, their methodology is currently state-of-the-art in the sphere of goal-object-oriented scenarios. We trained and evaluated both approaches on identical hardware and the same set of evaluation scenes to ensure an equitable comparison. The evaluation metrics we utilized are the same as those previously employed by ~\cite{xu2021efficient,fujita2020important}:

\begin{itemize}
    \item \textbf{Completion (C)}: The mean percentage completion over $n$ test runs. Completions are successful and equal to $1$ in a test run; if the system does not exceed in failing to grasp the goal object $n=5$ times, it is $0$. The metric measures the system's ability to complete the task.

    \item \textbf{Grasp success (GS)}: The mean percentage of successful grasp over all grasp attempts. This metric represents the accuracy of the model and its ability to estimate the goal object successfully grasped.
    \item \textbf{Motion number (MN)}: The mean number of push actions per completion. It reflects action efficiency.
\end{itemize}

\subsubsection{Sim-to-Sim}
CoppeliaSim is well-known for robust robotic testing due to its versatile models and interface~\cite{chen2023maniware, zhao2020sim}. In contrast, Gazebo closely mirrors real-world conditions with its advanced physics~\cite{kleeberger2020survey, bellegarda2022robust}. While CoppeliaSim excels in grasping simulations, Gazebo, relying on the Grasp-fix-plugin~\cite{wang2021rascal}, sometimes inaccurately executes grasping~\cite{hanna2021grounded}. Our Sim-to-Sim method trains in CoppeliaSim and fine-tuned in Gazebo (Fig.~\ref{fig:packedExps}), merging the strengths of both for optimal accuracy.

\begin{figure*}[!t]
    \includegraphics[width=0.245\linewidth]{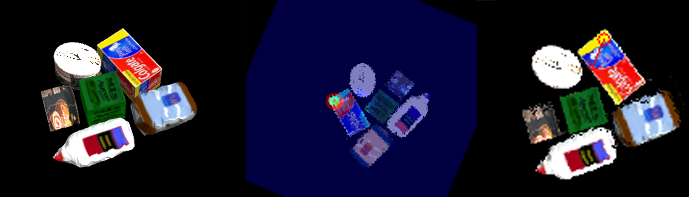}
    \includegraphics[width=0.245\linewidth]{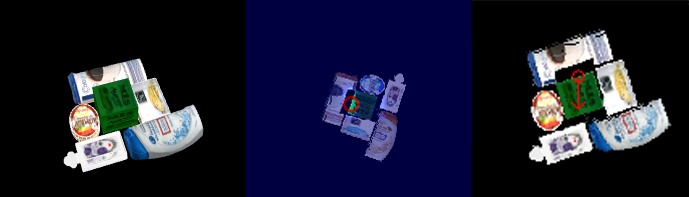}
    \includegraphics[width=0.245\linewidth]{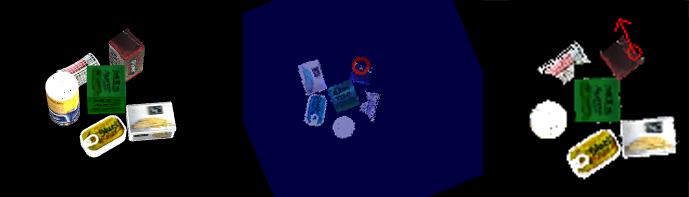}
    \includegraphics[width=0.245\linewidth]{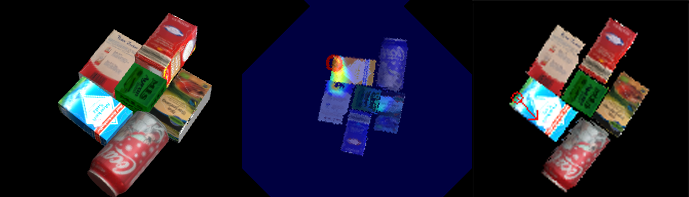}
    
    \par\noindent\makebox[0.245\linewidth][c]{\footnotesize(case:11)} \hfill
    \makebox[0.245\linewidth][c]{\footnotesize(case:12)} \hfill
    \makebox[0.245\linewidth][c]{\footnotesize(case:13)} \hfill
    \makebox[0.245\linewidth][c]{\footnotesize(case:14)} 

\caption{Four simulated scenes with novel, untrained objects in dense household clutter. The target in each is the green box. The initial column presents the original image; the middle one is the model's angle and position; the final is the action.}
\label{fig:6}
\end{figure*}

\subsection{Real Robot Experiments}
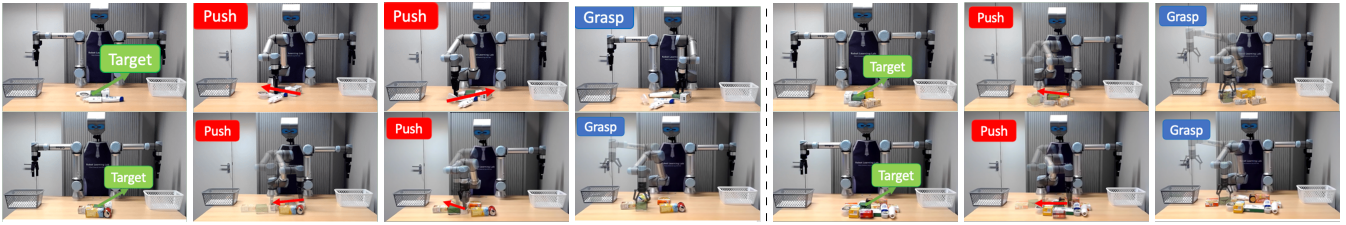
\begin{figure*}[!t]
    \includegraphics[width=0.138\linewidth]{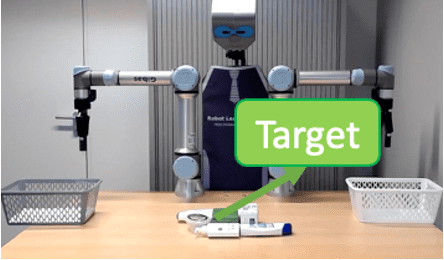}
    \includegraphics[width=0.138\linewidth]{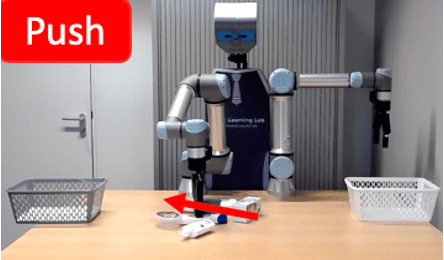}
    \includegraphics[width=0.138\linewidth]{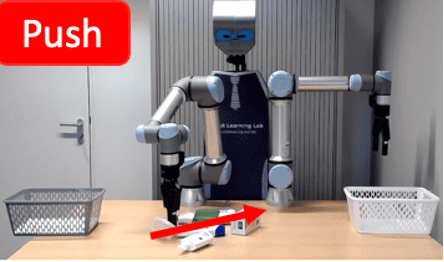}
    \includegraphics[width=0.138\linewidth]{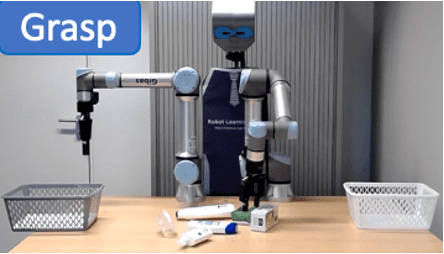}
    \begin{tikzpicture}
        \draw[dashed, line width = 0.01pt] (0,0) -- (0,0.08\linewidth);
    \end{tikzpicture}
    \includegraphics[width=0.138\linewidth]{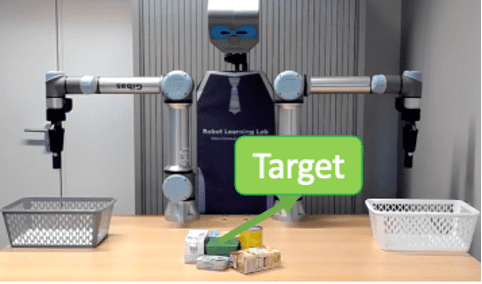}
    \includegraphics[width=0.138\linewidth]{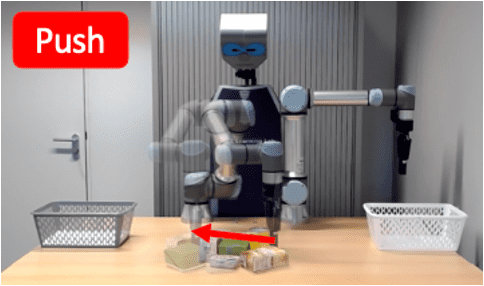}
    \includegraphics[width=0.138\linewidth]{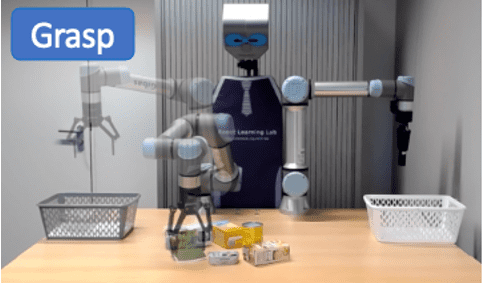}
    \includegraphics[width=0.138\linewidth]{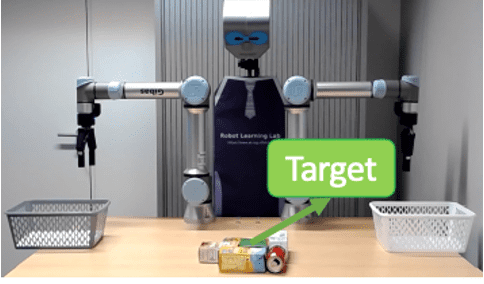}
    \includegraphics[width=0.138\linewidth]{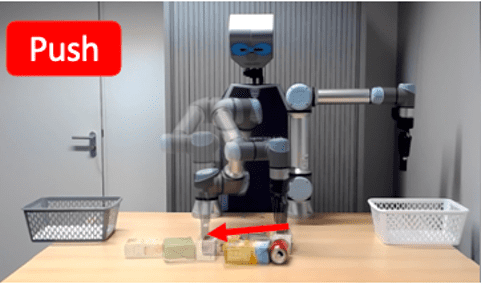}
    \includegraphics[width=0.138\linewidth]{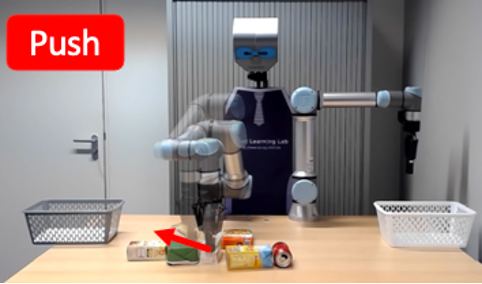}
    \includegraphics[width=0.138\linewidth]{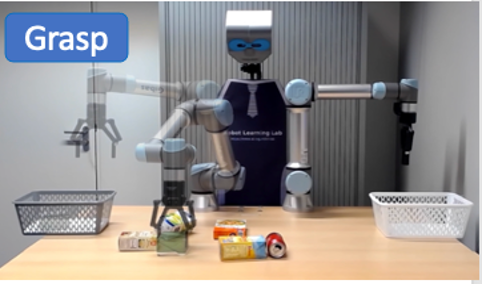}    
    \begin{tikzpicture}
        \draw[dashed, line width = 0.01pt] (0,0) -- (0,0.08\linewidth);
    \end{tikzpicture}
    \includegraphics[width=0.138\linewidth]{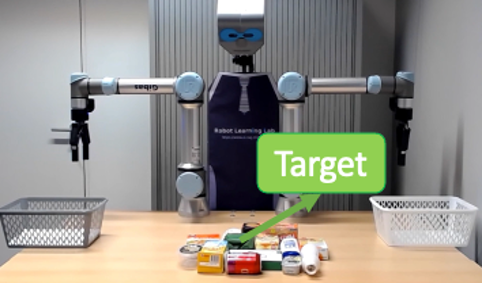}
    \includegraphics[width=0.138\linewidth]{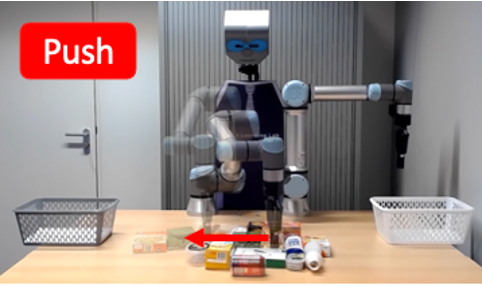}
    \includegraphics[width=0.138\linewidth]{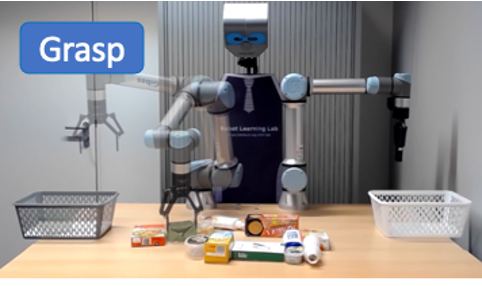}
    \caption{Real-world Experiments: Four untrained objects in scenes where clutter obstructs direct goal grasping.}
    \label{fig:real_robot}
\end{figure*}

We verified our methodology through real-world testing experiments using robots and objects that mirror our simulations. In untrained and highly cluttered environments and amid household items, our method, as depicted in Fig.~\ref{fig:real_robot}, showcased performance akin to human-like cognition, emphasizing its viability for kitchen service robots.

\subsection{Results Analysis}
We conducted four experiments to assess our approach. In Coppeliasim, we evaluated densely arranged and randomized object scenes. Subsequently, we used Gazebo to examine household objects. The final test gauged a physical robot's performance with a target object in multifaceted sequences, echoing real-world challenges.

\subsubsection{Packed scenes}
In the initial experiment, we assessed 9 intricate scenarios, each iterated 100 times. Results and their heat maps are found in Fig.~\ref{fig:5} (case:1-9) and Table~\ref{table:res1}. The minimal standard error showcases the method's reliable action sequence. The model appears to value push rewards (see Eq. \ref{eq:preward}), indicating efficient surrounding object clearance by relocating the target. Elevated $Q$-values adjacent to the target in Fig.~\ref{fig:packedExps} emphasize that creating space facilitates grasping. From the illustrative examples in Fig.~\ref{fig:5}, we posit that a single push is often sufficient to effectively grasp the target object.

In packed scenes, we choose variously shaped target objects, consistently arranged from scenes case:1 to case:9, with 5-12 objects each. Table~\ref{table:res1} shows that our method reaches an impressive completion rate of over $95\%$, outdoing Xu et al.~\cite{xu2021efficient} by $16\%$. Moreover, we exceed their Grasp Success (GS) by approximately $60\%$. Xu et al. struggle due to inaccurate space estimation during training and inference, primarily from ineffective masking, causing early experiment conclusions. In contrast, our model accurately discerns optimal grasps, leading to greater success. In scenes (case:2, case:3, and case:5), the cylindrical object's tendency to roll poses challenges for both methods, as detailed in Table~\ref{table:res1}. However, our method, with its subtle push strategy, ensures stability and outpaces Xu et al.~\cite{xu2021efficient}, showcasing its real-world applicability.

\subsubsection{Random Scenes}
In our second experiment, we assess scenes with 10, 15, and 20 objects, evaluating our method on 100 scenes for each count. As shown in Fig.~\ref{fig:5} case:10, object layouts are randomly generated, sometimes hiding the goal object. The results in Table~\ref{table:res2} raise questions about the goal mask's necessity. Though models with and without the mask show comparable completion rates, the latter needs an extra action. However, its reduced size makes it an appealing alternative. Although our approach takes more actions in 10-object scenes, it ensures higher success rates. While adding objects initially does not impact performance, complexities in 20-object scenarios reduce efficiency. However, our method consistently outperforms Xu et al.~\cite{xu2021efficient}, especially in 20-object scenes with a $20\%$ higher success rate. Despite having a higher mean number of actions (MN) than Xu et al.~\cite{xu2021efficient}, the robustness of our method is evident across varying scene complexities.

\subsubsection{Household Scenes}
In our household object experiments, we utilize the Gazebo simulator to evaluate our approach on 4 untrained simulated household items, always targeting to grasp a green milk box (see Fig.~\ref{fig:6}). The design of scenes mimics real-world challenges in complex environments. Based on the data in Table~\ref{table:res3}, after conducting 100 repetitions for each scenario, our approach consistently yields not only a superior GS but also maintains a completion rate of $100\%$.

\subsubsection{Real Robot Scenes}
As shown in Fig.~\ref{fig:real_robot}, our strategy enables the robot to adeptly grasp a target in real-world cluttered environments using our devised method. The figure underscores that, even in densely populated settings such as the fourth environment, our method often accomplishes the task with a mere 1-2 pushes. For a more comprehensive view, please refer to the accompanying video.

\begin{table}[!t]
    \begin{center}
        \caption{Simulation results for 9 complex packed scenes with 100 trials per scene}
        \resizebox{\linewidth}{!}{
        \begin{tabular}{| c | c | c | c | c |}
        \hline
        Approach & \#scene &\textbf{C\%} & \textbf{GS\%} & \textbf{MN} \\
        \hline
         & case:1 & $83.0 \pm 3.75$ & $26.48 \pm 2.16$ & $3.43 \pm 0.31$ \\
         & case:2 & $97.98\pm 1.42$ & $51.82\pm 3.63$ & $\mathbf{2.78\pm 0.29}$ \\ 
         & case:3 & $100.0\pm 0.0$ & $58.96\pm 3.75$ & $\mathbf{5.64\pm 0.39}$ \\ 
         & case:4 & $99.0\pm 0.99$ & $67.11\pm 3.88$ & $\mathbf{3.53\pm 0.47}$ \\ 
        Xu et al.~\cite{xu2021efficient} & case:5 & $\mathbf{100.0\pm 0.0}$ & $60.89\pm 3.66$ & $\mathbf{5.33\pm 0.52}$ \\ 
         & case:6 & $98.0\pm 1.41$ & $57.47\pm 3.77$ & $\mathbf{11.36\pm 1.54}$ \\ 
         & case:7 & $96.0\pm 1.97$ & $44.84\pm 3.32$ & $\mathbf{2.44\pm 0.32}$ \\ 
         & case:8 & $95.0\pm 2.19$ & $46.95\pm 3.41$ & $11.1\pm 1.26$ \\
         & case:9 & $100.0 \pm 0.0$ & $68.49\pm  3.86$ & $4.49\pm  0.32$  \\\hline
        
         & case:1 & $\mathbf{98.98 \pm 1.01}$ $\cred{\uparrow}$  & $\mathbf{86.08 \pm 3.32}$ $\cred{\uparrow}$ & $\mathbf{1.12 \pm 0.03}$ $\cred{\uparrow}$  \\ 
         & case:2 & $\mathbf{99.0 \pm 0.99}$ $\cred{\uparrow}$ & $\mathbf{66.67\pm  3.88}$ $\cred{\uparrow}$ & $4.99\pm  0.43$  \\ 
         & case:3 & $\mathbf{100.0 \pm 0.0}$ $\cred{\uparrow}$  & $\mathbf{77.52 \pm 3.69}$ $\cred{\uparrow}$ & $5.44 \pm 0.31$  \\ 
         & case:4 & $\mathbf{100.0 \pm 0.0}$ $\cred{\uparrow}$ & $\mathbf{83.33\pm  5.44}$ $\cred{\uparrow}$ & $5.88\pm  0.46$  \\ 
        Ours & case:5 & $98.98 \pm 1.02$  & $\mathbf{70.5 \pm 3.91}$ $\cred{\uparrow}$ & $8.97\pm  0.72$  \\ 
         & case:6 & $\mathbf{99.0 \pm 0.99}$ $\cred{\uparrow}$ & $\mathbf{86.21 \pm 3.3}$ $\cred{\uparrow}$ & $17.67 \pm 1.50$  \\ 
         & case:7 & $\mathbf{100 \pm 0.0}$ $\cred{\uparrow}$  & $\mathbf{61.73\pm  3.83}$ $\cred{\uparrow}$ & $3.04 \pm 0.14$  \\ 
         & case:8 & $\mathbf{99.0 \pm 1.0}$ $\cred{\uparrow}$ & $\mathbf{62.89\pm  3.86}$ $\cred{\uparrow}$ & $\mathbf{2.41\pm  0.089}$ $\cred{\uparrow}$  \\
         & case:9 & $\mathbf{100.0 \pm 0.0}$ $\cred{\uparrow}$ & $\mathbf{91.74\pm  2.65}$ $\cred{\uparrow}$ & $14.34\pm  1.60$  \\\hline

    \end{tabular}}
    \label{table:res1}
    \end{center}
\end{table}

\begin{table}[!t]
    \begin{center}
        \caption{Simulation results for random objects scenes with 100 trials per scene}
        \resizebox{\linewidth}{!}{
        \begin{tabular}{| c | c | c | c | c |}
        \hline
        Approach & \#objects &\textbf{C\%} & \textbf{GS\%} & \textbf{MN} \\
        \hline
         & 10 & $71.56\pm 4.48$ & $20.35\pm 1.57$ & $\mathbf{0.64\pm 0.25}$ \\
        Xu et al.~\cite{xu2021efficient} & 15 & $71.0\pm 4.56$ & $22.72\pm 1.75$ & $\mathbf{1.1\pm 0.54}$ \\ 
         & 20 & $77.89\pm 4.27$ & $21.59\pm 1.78$ & $\mathbf{1.13\pm 0.45}$ \\ \hline
        
         & 10 & $\mathbf{98.97\pm1.02}$ $\cred{\uparrow}$  & $66.21\pm  3.91$ & $1.02\pm  0.14$  \\ 
        Ours (mask) & 15 & $\mathbf{100 \pm 0.0}$ $\cred{\uparrow}$& $\mathbf{74.60\pm  3.89}$ $\cred{\uparrow}$ & $3.67\pm  0.98$  \\ 
         & 20 & $97.22\pm 1.95$  & $\mathbf{69.23\pm  4.62}$ $\cred{\uparrow}$ & $6.09\pm  1.82$  \\ \hline

         & 10 & $\mathbf{98.78\pm1.22}$ $\cred{\uparrow}$  & $\mathbf{68.33\pm  4.29}$ $\cred{\uparrow}$ & $3.19\pm  0.56$  \\ 
        Ours (no mask) & 15 & $\mathbf{100 \pm 0.0}$ $\cred{\uparrow}$ & $71.42\pm  4.28$ & $3.27\pm  0.47$  \\ 
         & 20 & $\mathbf{99.22\pm 0.1}$ $\cred{\uparrow}$  & $67.36\pm  4.83$ & $7.12\pm  2.04$  \\ \hline
    \end{tabular}}
    \label{table:res2}
    \end{center}
\end{table}

\begin{table}[!t]
    \begin{center}
        \caption{Simulation results for 4 household scenes with 100 trials per scene}
        \resizebox{\linewidth}{!}{
        \begin{tabular}{| c | c | c | c | c |}
        \hline
        Approach & \#scene &\textbf{C\%} & \textbf{GS\%} & \textbf{MN} \\
        \hline
         & case:11 & $95.96\pm 1.99$ & $46.48\pm 3.41$ & $3.75\pm 0.88$ \\ 
        Xu et al.~\cite{xu2021efficient} & case:12 & $94.0\pm 2.39$ & $41.67\pm 3.16$ & $2.58\pm 0.46$ \\ 
         & case:13 & $97.0\pm 1.71$ & $57.80\pm 3.78$ & $\mathbf{1.73\pm 0.12}$ \\ 
         & case:14 & $99.0\pm 0.99$ & $58.82\pm 3.79$ & $\mathbf{1.47\pm 0.28}$ \\ \hline
        
         & case:11 & $\mathbf{100.0\pm0.0}$ $\cred{\uparrow}$ & $\mathbf{56.18\pm 3.73}$ $\cred{\uparrow}$ & $\mathbf{3.33\pm 0.29}$ $\cred{\uparrow}$ \\ 
        Ours & case:12 & $\mathbf{100.0\pm 0.0}$ $\cred{\uparrow}$ & $\mathbf{68.97\pm 3.86}$ $\cred{\uparrow}$ & $\mathbf{1.26\pm 0.05}$ $\cred{\uparrow}$ \\ 
         & case:13 & $\mathbf{100.0\pm 0.0}$ $\cred{\uparrow}$ & $\mathbf{75.76\pm 3.74}$ $\cred{\uparrow}$ & $2.49\pm 0.10$ \\
         & case:14 & $\mathbf{100.0\pm 0.0}$ $\cred{\uparrow}$ & $\mathbf{76.92\pm 3.71}$ $\cred{\uparrow}$ & $3.69\pm 0.21$ \\ \hline 
        \end{tabular}}
    \label{table:res3}
    \end{center}
\end{table}

\section{Conclusion and Future Work}
In this paper, we propose an innovative self-supervised DRL approach that enables robots to grasp target objects in highly cluttered and untrained environments. We conduct a comprehensive evaluation of our method across a spectrum of challenging scenarios, including densely packed arrangements of building blocks, random environments with varying object counts (10, 15, and 20 objects), and real-world object manipulation tasks. Notably, our approach achieves superior performance while maintaining a significantly smaller model size.

We design two distinct scenarios using building blocks, enhancing complexity by altering the shape of the target object. The results consistently demonstrate our method's proficiency in interacting with its surroundings and successfully grasping the target object. Overall, our performance in terms of task completion and grasp success surpasses Xu et al.'s, with minor variations in the Mean Number of Actions (MN) metric, primarily attributable to occasional misjudgments in their model. In the third experimental phase, we transition to real-world object manipulation tasks, where our approach consistently outperforms others, especially for some untrained objects in environments. To ensure a reliable evaluation, we employ a Sim-to-Sim strategy to mitigate inaccuracies introduced by Gazebo, ensuring a more trustworthy assessment. Impressively, our approach maintains its robustness and effectiveness in real-world testing, mirroring its strong performance in simulation. Our future research explores curriculum learning to further enhance our system's object manipulation proficiency.


\newpage
\bibliographystyle{IEEEtran}
\bibliography{main}
\newpage

\end{document}